\documentclass[runningheads]{llncs}
\usepackage[T1]{fontenc}
\usepackage{graphicx}
\PassOptionsToPackage{hyphens}{url}
\usepackage[colorlinks=true,
            citecolor=blue,
            linkcolor=blue,
            urlcolor=blue]{hyperref}

\usepackage[utf8]{inputenc}
\usepackage{amsmath,amssymb}
\usepackage[per-mode=fraction]{siunitx}
\providecommand{\qty}[2]{\SI{#1}{#2}}
\usepackage[caption=false]{subfig}
\usepackage[capitalize]{cleveref}
\usepackage{todonotes}
\usepackage{xspace}
\usepackage{xcolor}

\newcommand\ie{\mbox{i.\,e.}\xspace}
\newcommand\eg{\mbox{e.\,g.}\xspace}
\newcommand\cf{\mbox{cf.}\xspace}

\begin{document}
\title{B-Human '26 -- Scalable Team Play up to 11\,vs.\,11}
\titlerunning{B-Human '26 -- Scalable Team Play up to 11~vs.~11}
\author{Arne Hasselbring\inst{1}, Tim Laue\inst{1}, Thomas Röfer\inst{2,1}, Philip Reichenberg\inst{1}, }
\authorrunning{A. Hasselbring et al.}
\institute{Universität Bremen, Fachbereich 3 -- Mathematik und Informatik,\\
	Postfach 330 440, 28334 Bremen, Germany
	\and
	Deutsches Forschungszentrum für Künstliche Intelligenz,\\
	Cyber-Physical Systems, Enrique-Schmidt-Str. 5, 28359 Bremen, Germany\\
	 \email{\{arha,tlaue,p.reichenberg\}@uni-bremen.de\qquad thomas.roefer@dfki.de}}
\maketitle              % typeset the header of the contribution

\begin{abstract}
B-Human won multiple competitions in the Middle Division of the new Humanoid Soccer League in 2026.
In addition to regular matches with three or five players per team, we ventured into uncharted territory by playing demo matches with eleven humanoid robots, highlighting the tactical flexibility of our behavior and the overall robustness of our system.
As B-Human formerly competed in the Standard Platform League with robots half the size and less than a third the weight, we begin by outlining the adjustments needed for this transition.
The central section of this paper details our behavior architecture, drawing on concepts from RoboCup leagues that have used 11 players for much longer.
\end{abstract}

\section{Introduction}
\begin{figure}[t]
    \centering
    \includegraphics[width=0.8\textwidth]{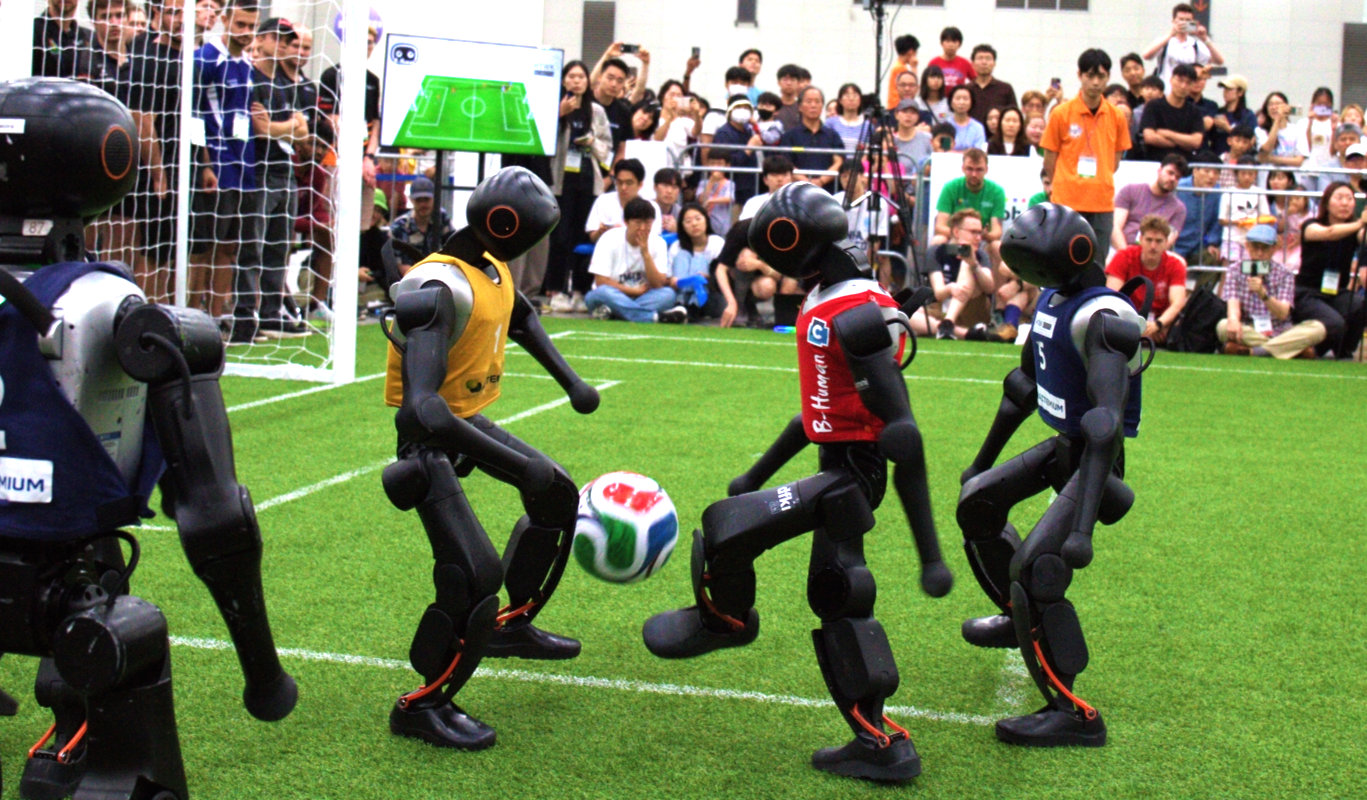}
    \caption{Booster K1 robots in action during RoboCup 2026}
    \label{f:robots}
\end{figure}

The RoboCup 2026 competition year marks a new milestone with the merge of the Standard Platform League (SPL) and the Humanoid League (HL) into the newly formed Humanoid Soccer League (HSL). This merger coincided with the availability of a wide variety of new, high-performance humanoid robots that are also eligible to compete in the HSL, divided into different divisions based on size and weight. Given this new league structure and different robots to choose from, we opted to play with the Booster K1 (see \cref{f:robots}) robot in the Middle Division instead of continuing to work with the NAO robot. This decision was supported by the fact that we already started to work with the Booster T1 robot last year \cite{Winner-RC-2025}, which made the transition toward the new robot platform easier.

An additional major challenge was adapting the system to the new set of rules resulting from the merger of the previous leagues. In addition to numerous minor changes, this also required the redevelopment of certain components, such as those for detecting robots and the ball (\cf \cref{ss:objectDetection}). One important new feature of the HSL concerns the parameters regarding field size and the number of players. The rules define three different field layouts, referred to as S-Field, M-Field and L-Field, with sizes ranging from \SI{54}{\square\meter} to \SI{308}{\square\meter}. At RoboCup 2026 as well as at the RoboCup German Open 2026, the games of the Middle Division were either played 3~vs.~3 on an S-Field or 5~vs.~5 on an M-Field. In addition, B-Human and HTWK Robots played the world's first 11~vs.~11 humanoid match on an L-Field during RoboCup 2026. Following this competition, B-Human also participated in the RoboCup Asia-Pacific (RCAP) Beijing Masters and the World Humanoid Robot Games (WHRG), where all matches were played 5~vs.~5 on L-Fields and another 11~vs.~11 demo match was carried out, this time against HTWhIRLK.
Given this number of possible configurations and the additional fact that individual robots are sometimes off the field -- for instance, due to time penalties -- a system is required that is always capable of successfully playing soccer with a team ranging in size from 1 to 11.

The main contribution of this paper is a description of our behavior approach (\cf \cref{s:behavior}), which is the key component to address the challenges described previously. Coordinating a large number of agents is not new in RoboCup and has already been realized in other leagues, namely the Simulation Leagues as well as the Small-Size League. Therefore, we have adopted some approaches from these leagues and applied them to humanoid robots, in line with the general idea of RoboCup. The behavior description is preceded by a brief overview of the system's other major components (\cf \cref{s:overview}) that provide the necessary information and available actions for the behavior. Finally, a compilation of all B-Human results in 2026 Middle Division competitions is presented in \cref{s:results}.

\section{System Overview}
\label{s:overview}
With the switch to the new hardware platform, B-Human did not start from scratch; instead, we continued to use -- or rather, expanded upon -- the system previously employed for the NAO. It has a modular design and consists of well over 100 components, each of which solves a subproblem of the overall task of playing soccer. Some of these implementations, such as those for state estimation (\cf \cref{ss:stateEstimation}), could be kept with only minor modifications, while others, such as the detection for the ball and other robots (\cf \cref{ss:objectDetection}) or the motion implementation (\cf \cref{ss:rl}), required completely new solutions. This section provides a brief overview of the current state and the applied algorithms and approaches of the most important components.

\subsection{Object Detection}
\label{ss:objectDetection}

\begin{figure}[t]
    \centering
    \includegraphics[height=4.9cm]{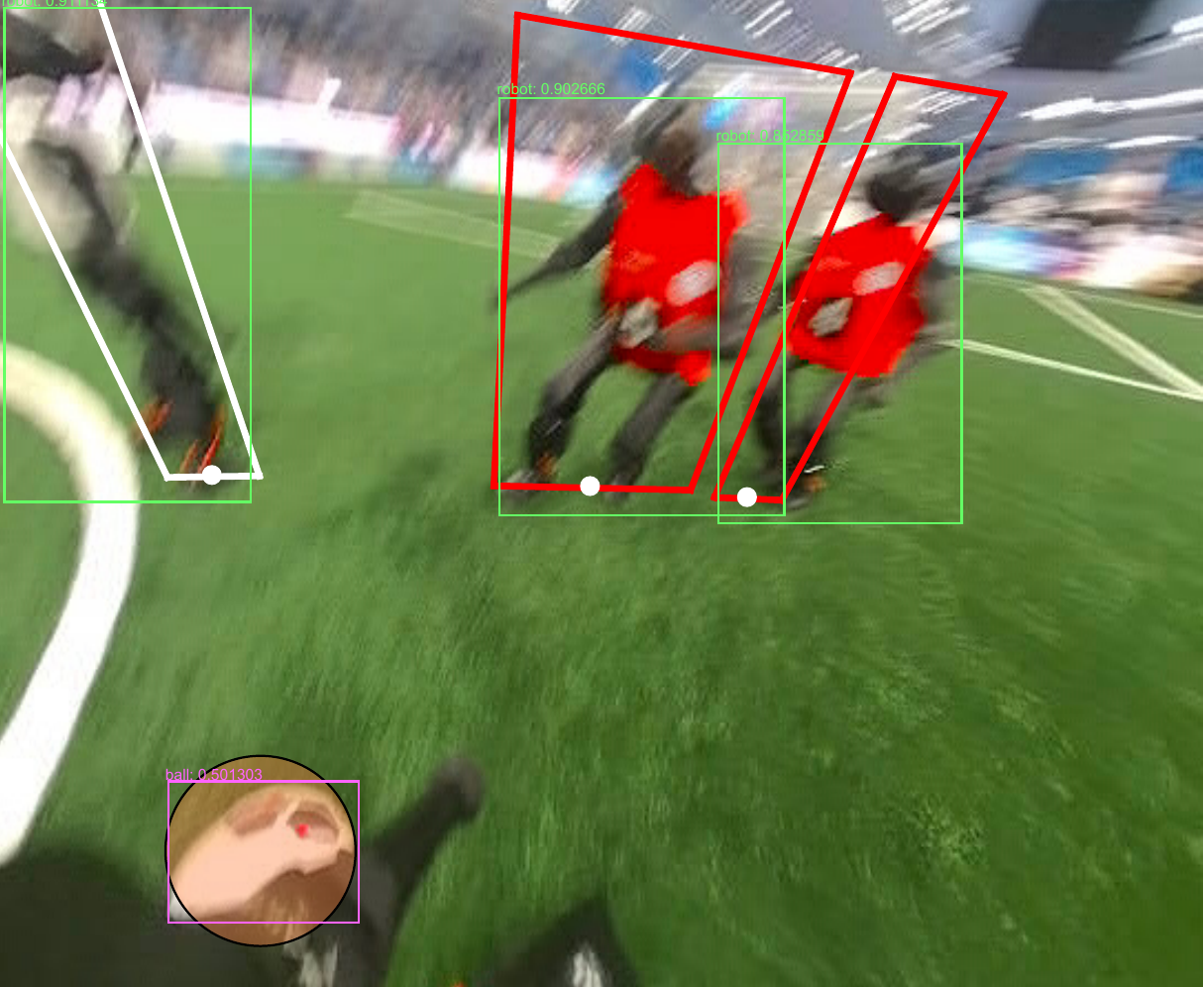}\hfill
    \includegraphics[height=4.9cm]{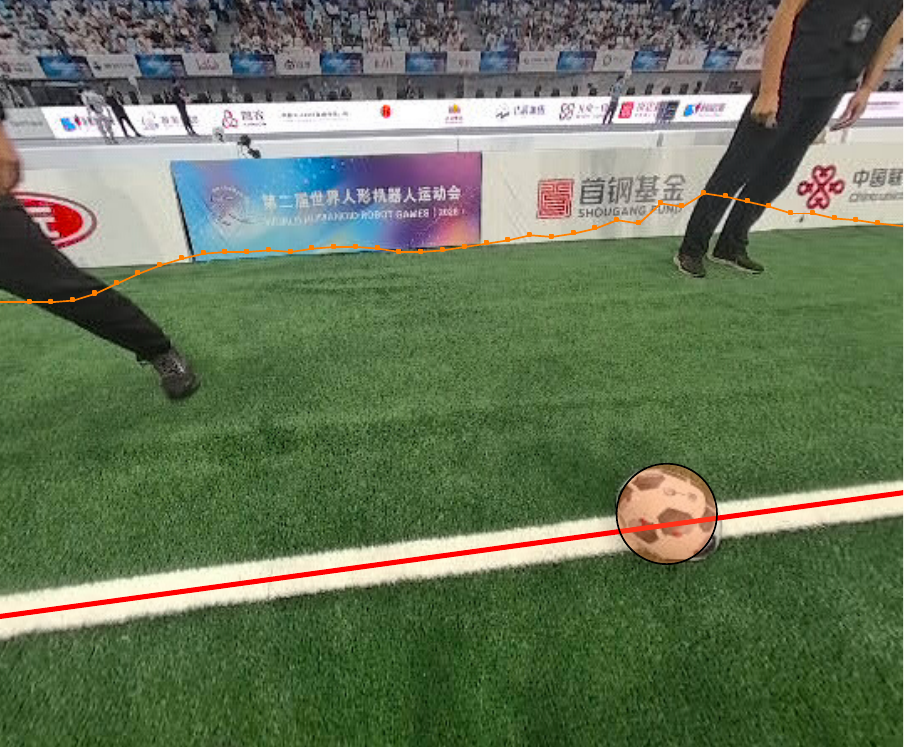}
    \caption{\label{f:objectDetection}Two images with marked recognized objects from the perspective of a robot during a game. Left: The violet and green rectangles show the output of the YOLOX detector. The quadrilaterals, which each have the color of the recognized jersey, show the perspective-corrected representation. The orange circle denotes the refined ball position. Right: Exemplary detection of a touchline.}
\end{figure}

Two primary distinctions between the Standard Platform League and the Humanoid Soccer League are the appearance of the ball and the players. Consequently, we trained a YOLOX~\cite{yolox2021} network to detect balls and humanoid robots. It runs at \qty{60}{\hertz} and the dataset can be found on our website\footnote{\url{https://www.b-human.de/datasets-en.html}}. For training, we utilized the TORSO-21~\cite{RC-Bestmann-etal-2022} dataset, which was supplemented by numerous images captured by our robots, which we manually labeled. However, due to the wide field of view of the Booster K1's camera, the bounding boxes provided by the YOLOX network are not directly suitable for determining the location of the ball or the robots, as depicted in the left part of \cref{f:objectDetection}. This is because the ball appears elongated near the image border, and the robots appear more diagonal as they move further left or right in the image. To address this issue, we employ an iterative approach to refine the position of the ball. For a more realistic estimate of the position and width of a detected robot, we apply a heuristic that -- given the camera parameters and an assumed height for the robot's center -- determines individual ground points more precisely.

The detection of all field elements, such as the lines, their intersections and the penalty spot, remains the same as it was for the NAO robot, a combination of algorithmic processing and the classification of small image sections. In the HSL, most teams currently seem to be focusing on detecting intersections rather than lines. However, we also consider lines to be extremely important, as they help to reduce the localization error significantly in many situations, such as the one shown in the right part of \cref{f:objectDetection}.

\subsection{State Estimation}
\label{ss:stateEstimation}
\begin{figure}[t]
    \centering
    \includegraphics[width=0.8\textwidth]{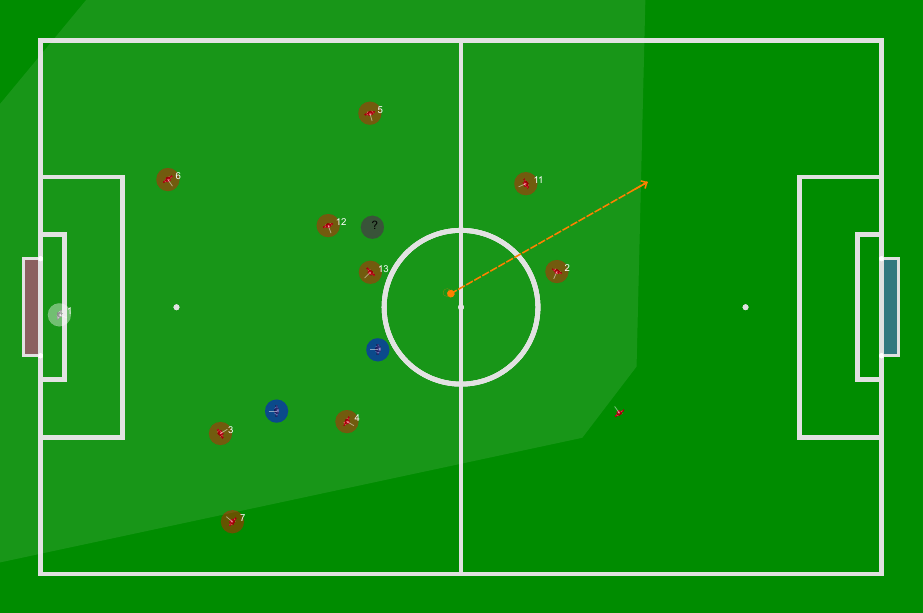}
    \caption{World model of a B-Human robot (the rightmost robot, depicted together with its current field of view) during the 11~vs.~11 match at RoboCup 2026. Red and white circles with numbers denote the positions of the teammates, blue and gray circles indicate opponents and unknown robots, respectively. The currently rolling ball is drawn in orange, along with an arrow that denotes the position at which it is assumed to stop.}
    \label{f:worldmodel}
\end{figure}

To play soccer effectively as a team -- which entails, among other things, sensible positioning and the execution of precise passes -- each robot requires a robust estimate of the current state of the world: specifically, its own position as well as the positions of the ball and the other robots, as depicted in \cref{f:worldmodel}. B-Human has possessed such a world model for several years and was therefore able to implement complex cooperative behavior using NAO robots, as described in \cite{Winner-RoboCup-2023}. Most of the components involved have remained in use without major changes. Only a few parameters had to be adjusted due to the changed sizes and speeds of the ball and the robots.

For self-localization, we still use a particle filter with only a small number of particles, each of which, however, contains an Unscented Kalman filter.
Detected field lines, line intersections, penalty spots, and the center circle serve as input, though not the goals, for which we currently do not have a reliable detection. A systematic evaluation of localization accuracy against an external tracking system has not been conducted. Informal inspection of recorded data alongside match videos, however, indicates that no large errors occur and that the typical deviation is below approximately \qty{10}{\centi\meter}.

For the ball state, we constantly maintain a set of different hypotheses (stationary as well as rolling), from which the most probable one is used. Each individual hypothesis employs a Kalman filter for state estimation. Together with a friction model, we use the estimated velocity to predict a rolling ball's trajectory, as depicted in \cref{f:worldmodel}. This is particularly useful for anticipatory robot behaviors.

To play successfully as a team, it is essential to know where your teammates are positioned. The estimation of their positions is based on the messages exchanged among them, which contain, among other things, positions, movement destinations, and information about the ball. The current HSL allows a huge number of messages to be sent within the team. However, as this might become reduced again in the future and experience shows that communication problems often occur in competitions, we kept our approach of fusing observations of teammates (\cf \cref{ss:objectDetection}) with the communicated positions and motions. This creates a complete and robust model of the team, as shown exemplarily in \cref{f:worldmodel}. Opponent information is currently not communicated. Thus, the robots operate on a model that only includes recently seen opponent robots. They are tracked using Kalman filters.

\subsection{Motion Skills}
\label{ss:rl}
Given the new robot platform and the resulting opportunity to apply state-of-the-art reinforcement learning to motion control, we decided to use it to learn all three basic motion skills required for the competition: walking, kicking with different strengths, and standing up after a fall. Our approach is described in detail in \cite{RL-Kicking}, and the code is publicly available at \cite{GitHub-BHuman-RLStandup}.

\section{Behavior}
\label{s:behavior}

\begin{figure}[t]
    \includegraphics[width=\textwidth]{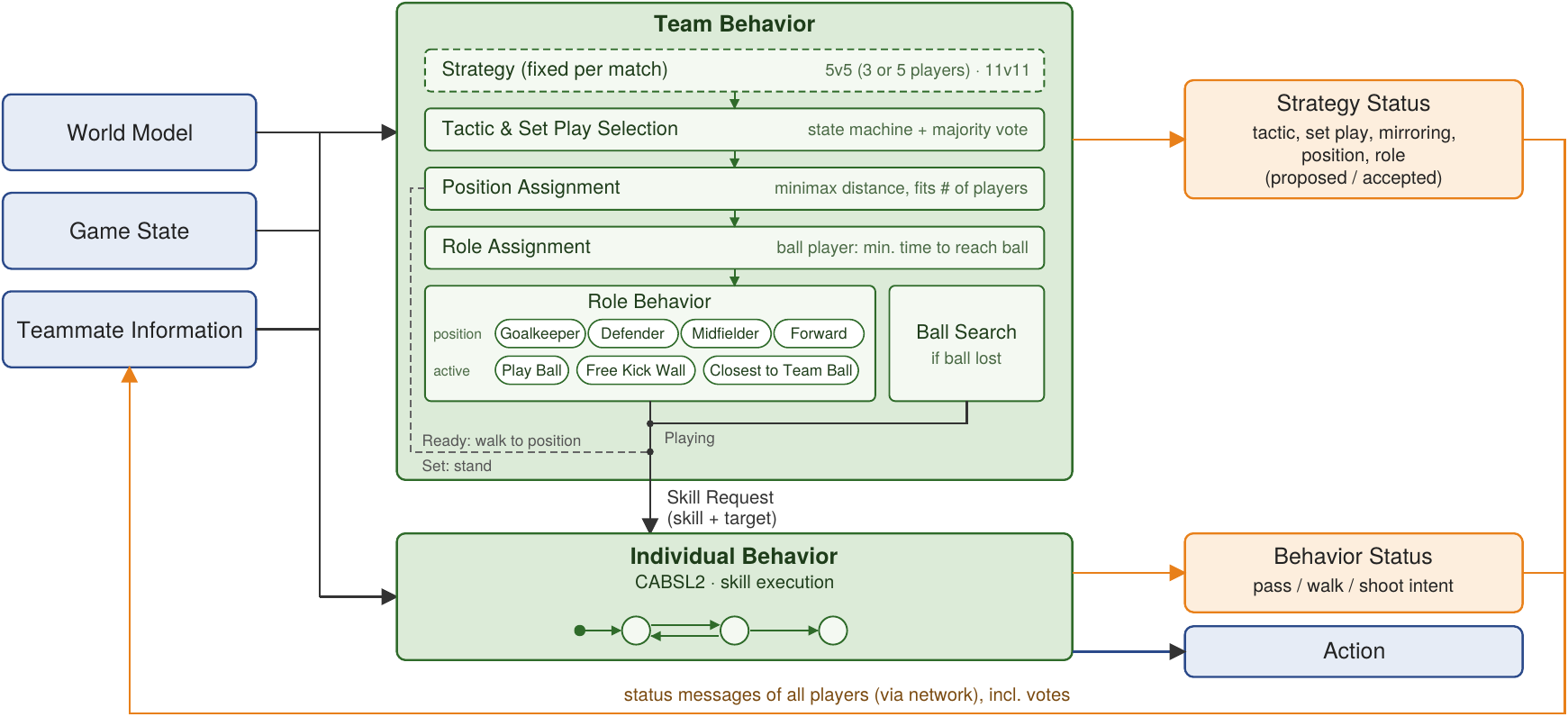}
    \caption{Architecture of the behavior}
    \label{f:behavior}
\end{figure}

Behavior is divided into two layers:
a partly configuration-based team behavior module, and a CABSL2-based~\cite{Roefer-RC-2025} module for individual skills.
The architecture, with a focus on team behavior, is shown in~\cref{f:behavior}; its components are explained in the following sections.
It bears a resemblance to systems used in the 2D Simulation League~\cite{Reis-2001} or Small Size League~\cite{Browning-JSCE-2005} over 20 years ago.

Team coordination is closely related to communication among the players, which B-Human does exclusively using wireless LAN.
The HSL limits the number of messages that can be sent per match to \num{12000} per team.
Although this is ten times as much as in the final years of the SPL for fewer players, for which the architecture was designed, we did not revert to an overly centralized coordination method.
The performance benefits from up-to-date data from teammates anyway.

\subsection{Strategy}
\label{ss:strategy}

\begin{figure}[t]
    \includegraphics[width=\textwidth]{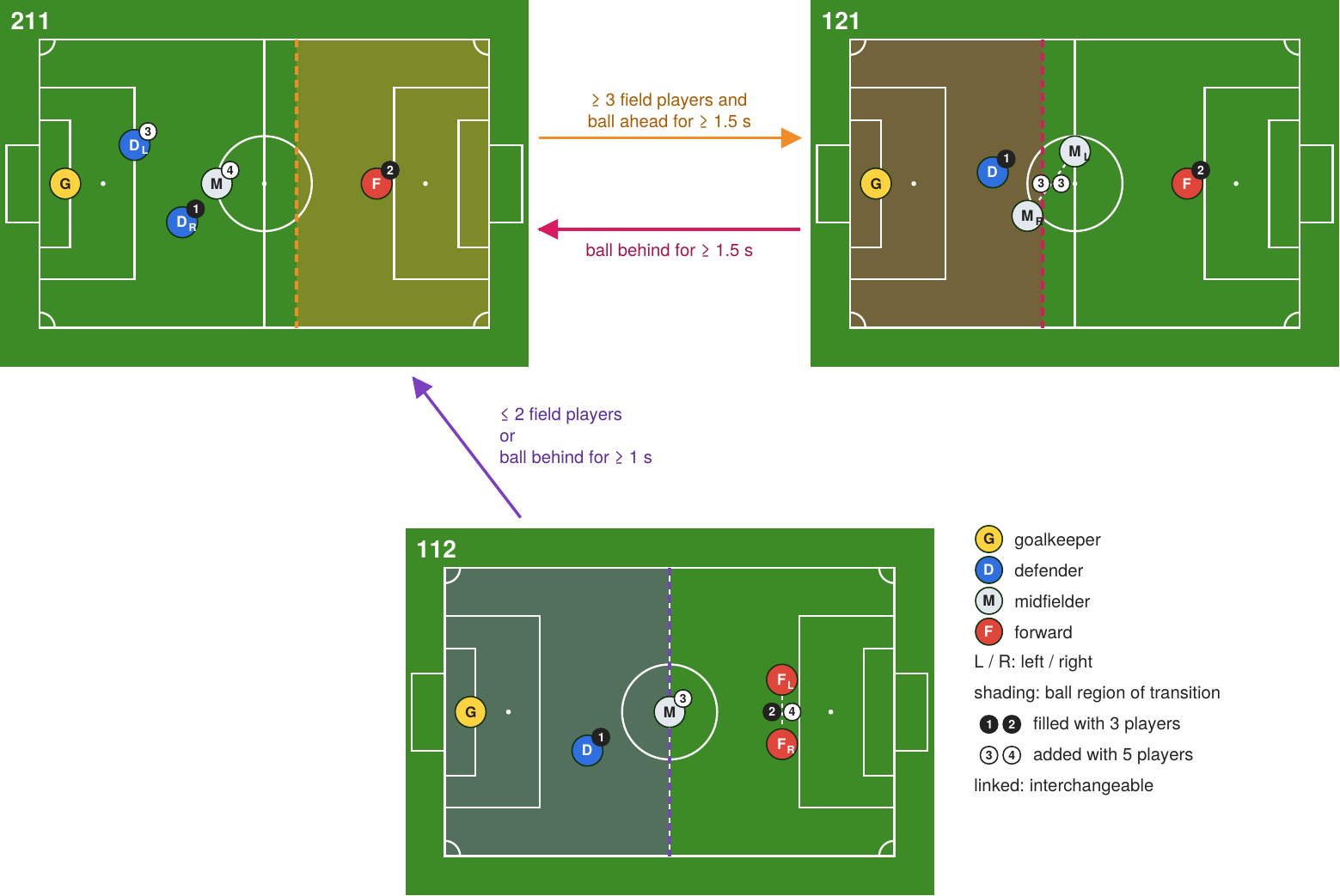}
    \caption{The strategy used at RoboCup 2026 for regular 3~vs.~3 and 5~vs.~5 matches.
Tactics can also be invoked by selecting a derived set play (not depicted here), thus \texttt{112} may be selected.
The priorities (lower $\widehat{=}$ more important) are represented by circled numbers at the positions.
Groups of numbers and positions connected by a dashed line mean that any of the positions may fill any of the priorities, \eg in the \texttt{112} tactic the forward in 3~vs.~3 may be either left or right.}
    \label{f:strategy5v5}
\end{figure}

At the top level, the team behavior is determined by a \emph{strategy} which is fixed for a match.
We currently have one strategy for up to five players (\texttt{5v5}, \cf~\cref{f:strategy5v5}) and one for up to eleven (\texttt{11v11}).
The strategy defines which \emph{tactics} are available, the conditions for transitioning between them, and lists of set plays for different game states, \ie kick-offs, penalty kicks, and free kicks.

A \emph{tactic} contains a list of positions, which are pairs of a label such as \texttt{midfielderL}, and locations on the field.
Our tactics are named, similar to popular football tactics, by the number of defenders, midfielders, and forwards, \eg \texttt{112} for an offensive tactic in the \texttt{5v5} strategy.
The label of a position selects the positioning algorithm (\cf~\cref{ss:positioning}).
The location is used in the assignment of positions (\cf~\cref{ss:roleAssignment}), as an anchor in the positioning algorithm, and to define a Voronoi diagram of ``areas of responsibility'' for each position (\cf~\cref{f:voronoiRegions}).
Each tactic further specifies which positions are left unoccupied if there are fewer players than expected, \eg due to penalties, or playing in the Foundation configuration~\cite[\S 3.1]{hsl_rulebook} with the \texttt{5v5} strategy.
The specification is a generalization of a prioritized list.
This allows multiple positions to have the same priority and defers the decision to the position assignment (\cf~\cref{ss:roleAssignment}), which chooses the cheaper subset of positions.
The \texttt{5v5} tactics and transitions are shown in~\cref{f:strategy5v5}.

\emph{Set plays} are based on a given tactic and override some positions by shifting their locations, while others stay at the default from the tactic.
They have preconditions that restrict the region where the ball may be.

Tactics and free kicks are selected by majority vote among all players of the team, \ie each player's team message contains its vote for a tactic / free kick.
Each player does the evaluation locally; ties are broken in favor of the option that the lower-numbered player voted for.
Currently, all tactic switches are triggered by the ball being ahead of or behind a certain boundary for enough time, and the number of available field players.
Kick-offs and penalty kicks are selected deterministically using data that all players know (such as the current score), which eliminates the need for communication during the \emph{ready} state.

\subsection{Position / Role Assignment}
\label{ss:roleAssignment}

After selecting a tactic or set play, the positions must be assigned to players.
Each player does so not only for itself but creates a hypothesis for the entire team.
First, the goalkeeper is a special case and is assigned given its player number, as selected in the GameController.
We do not replace an absent goalkeeper's position with a field player.
For the remaining positions, a cost matrix from players to positions is constructed with the distances between the players' last communicated locations and the positions' locations as specified in the tactic or set play.
Players use their own last communicated location even if they have access to the current estimate, so that (at least without packet loss) all players calculate the same cost.
An entry in the matrix is reduced if the respective assignment was previously assumed.
The entire tactic may be left/right-mirrored if that lowers the overall cost (see below).
Similarly, set plays are specified for a canonical side of the ball and are mirrored if the ball is on the other side.

For each of the valid sets of $n$ target positions (determined by the tactic or set play, \cf~\cref{ss:strategy}), an assignment to the $n$ field players is calculated using the approach from MacAlpine et~al.~\cite{MacAlpine:13robocup}.
They treat it as a stricter linear bottleneck assignment problem (LBAP) where distances between current and target positions define costs.
A standard LBAP minimizes the highest individual assignment cost.
In this stricter version, it minimizes the worst cost, then the second-worst among ties, then the third-worst, and so on.
A brute-force method sorts individual costs of each assignment in descending order and checks all $n!$ assignments to find the lexicographically smallest sequence, which is the one that belongs to the best assignment.
We originally used this approach because it remained fast enough for the up to six field players we had in the RoboCup Standard Platform League.
With ten players, however, it becomes too slow for our current \qty{60}{\hertz} frame rate.
On a desktop computer, we measured about \qty{150}{\milli\second} per solution.
To play with eleven players, we required a much faster method.

MacAlpine et~al.~\cite{MacAlpine-AAAI-2015} outline several methods to address the assignment problem with greater efficiency.
We apply the Hungarian method~\cite{Kuhn1955Hungarian} using ranked assignment costs.
These costs arise from sorting the original values from lowest to highest, then mapping each to its position in the sorted sequence; repeated values take the first index.
The resulting ranked costs become integers where only the bit at that index is set, ensuring each exceeds the sum of all smaller values.
As a result, lowering the largest individual costs aligns with reducing total costs, so the Hungarian method solves the LBAP under ranked costs.
With up to 11 players in soccer, $11^2$ bits fit inside a single 128-bit value that modern processors manage easily.
In 11~vs.~11 matches with ten field players assigned, the method is roughly 7000 times faster than brute force, keeping runtime well below \qty{1}{\milli\second}.

Once positions are assigned, each player is given a role.
There are two kinds of roles: \emph{position} roles and \emph{active} roles.
The position role follows from the label of the assigned position (goalkeeper, defender, midfielder, or forward) and selects the positioning algorithm (\cf~\cref{ss:positioning}).
An active role temporarily overrides the position role of a player.
Normally, one player gets the active role of playing the ball.
When the opponent has a free kick, it instead waits as a wall between the ball and our goal at a legal distance.
A player that has not seen the ball itself but knows the team ball may get the role \emph{closest to team ball}, which makes it go there to look for the ball, if it is considerably closer than the player that is currently playing the ball.

This distinction between position and active roles has one important consequence compared to pre-2022 B-Human behavior:
A player chasing the ball does not lose its place in the overall tactic, where it would have been replaced by another player.
This generally spaces the players out more, a prerequisite for making passing useful.

The assignment of the active role is decentralized, \ie each player has its own belief that is updated infrequently by incoming team messages.
Players are compared by their estimated time to reach a pose behind the ball, facing the opponent's goal.
This time is increased if a player has to get up, has recently lost the ball, or is the goalkeeper while the ball is far from the goal.
The player that currently plays the ball receives a bonus to prevent oscillating switches.
Each player transforms its own ball model into the frame of the other players, as communicated by them, which has the advantage of being reactive to movement of the ball at the cost of going through two uncertain transformations (the own and the other player's pose).
The system is tuned such to prefer that more than one player chases the ball rather than none, \eg players whose last communicated ball observation does not match the own are excluded from the decision, so that each of them may go to its own ball.

\subsection{Positioning}
\label{ss:positioning}

\begin{figure}[t]
    \includegraphics[width=\textwidth]{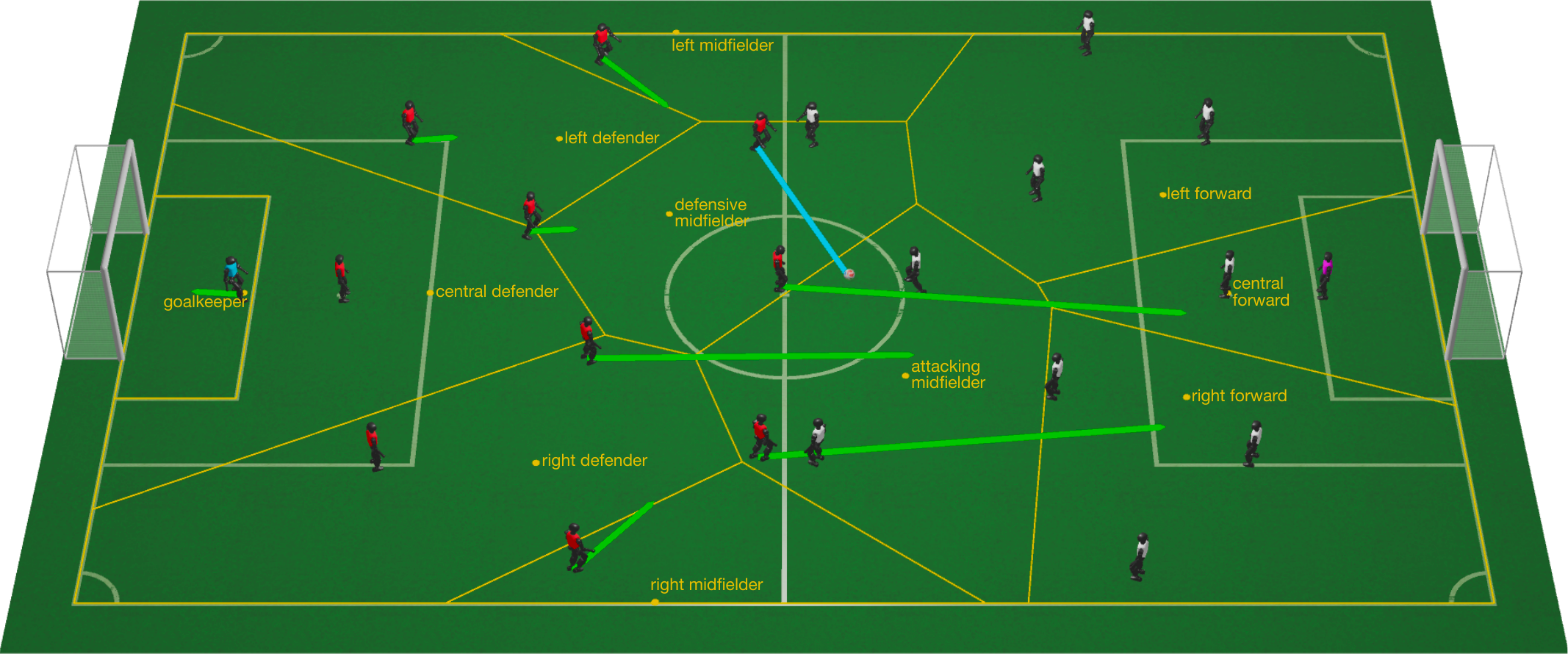}
    \caption{Base positions and corresponding Voronoi cells for the red team on the left, captured just after kick-off while the ball rolls from the center. Green arrows indicate the locations assigned to the red players. A cyan arrow identifies the player moving toward the ball rather than the left forward position, aiming to be the second to touch it because the HSL kick-off is indirect. The field is the one used at the RCAP Beijing Masters 2026.}
    \label{f:voronoiRegions}
\end{figure}

Players with a position role are positioned as described in~\cite{Winner-RoboCup-2023}:
Each player maximizes a role-specific rating function inside its own Voronoi cell (\cf~\cref{f:voronoiRegions}) using gradient ascent, starting from its previous result.
The cells are precomputed for each valid subset of positions of a tactic or set play (\cf~\cref{ss:strategy}), so a missing player simply enlarges the cells of its neighbors.

Besides the factors described there, the ratings penalize positions near the cell border, favor positions near the last communicated target, which keeps the behavior consistent with what the teammates assume, and exclude positions that would put the player on the other side of the line between the ball and the opponent's goal, so that it does not cross the path of the ball player.
The roles differ in their remaining factors:
The \emph{midfielder} shifts its anchor toward the ball when the ball is in the opponent's half to react quickly to counterattacks.
The \emph{defender} favors positions between the ball and our goal, marks opponents by standing between them and our goal (weighted by how dangerous their positions are), and avoids the corridor of a teammate's shot on goal.
At our own free kicks, defenders move away from the ball, whereas midfielders look for free space with a good scoring chance.
To prevent jittering, the previous result gets a bonus, targets are averaged over a short time, and a player only starts walking if the rating at the target is considerably better than at its current position.

%The goalkeeper does not use a rating function:
%It stands on the line between the ball and the goal at the distance where it covers the opening angle of the goal seen from the ball, but never in front of a base line at the goal area.

\subsection{Passing}
\label{ss:passing}

\begin{figure}[t]
    \centering
    \includegraphics[width=0.8\textwidth]{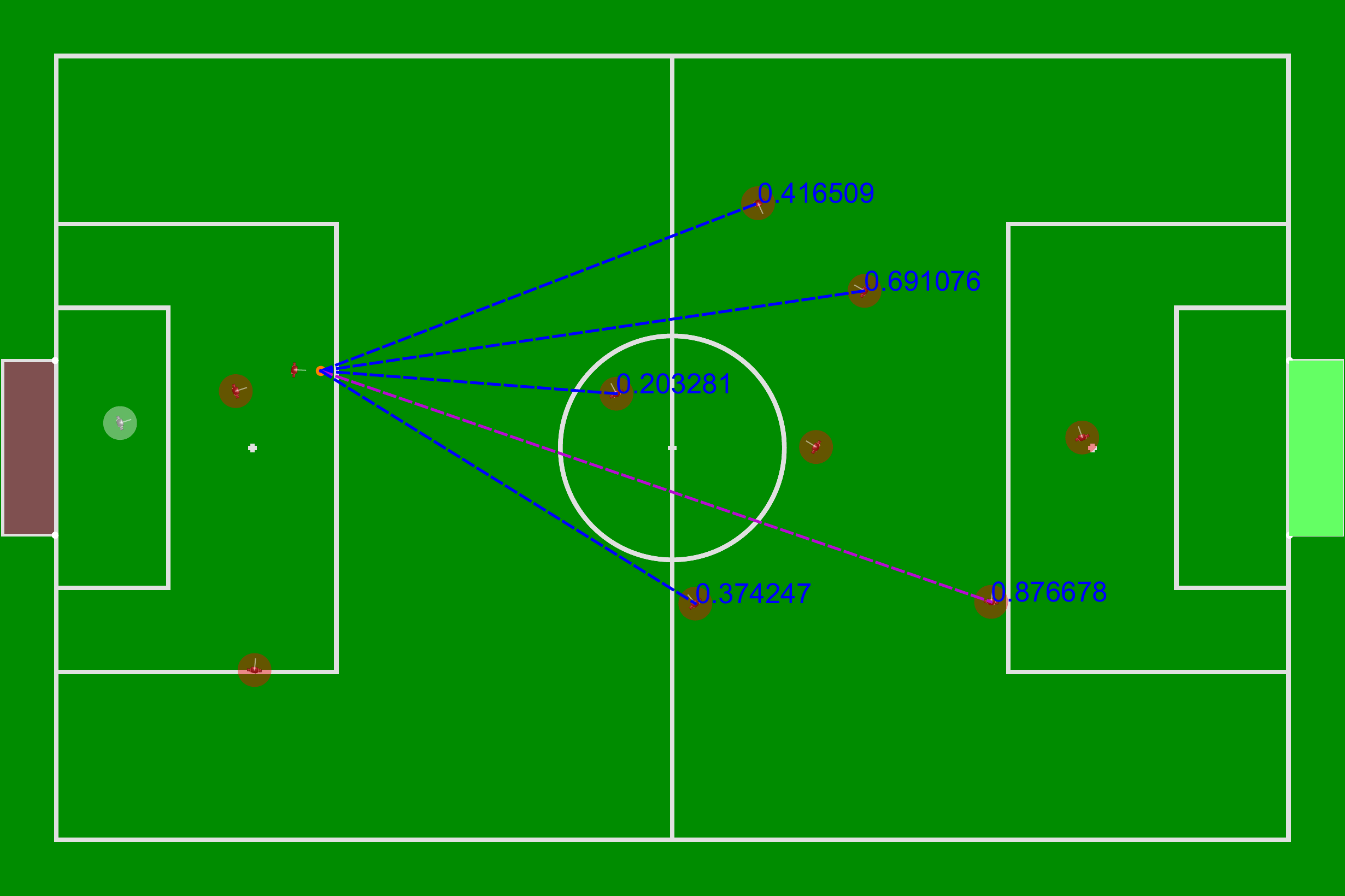}
    \caption{The combined rating to evaluate passes towards teammates. The blue line and number show the value of passes, which are higher rated than dribbling or trying to score a goal directly. The violet line shows the selected pass target.}
    \label{f:passing}
\end{figure}

Our passing involves teammates positioning for potential passes while the player with the ball selects the teammate most likely to score, as noted in~\cite{Winner-RoboCup-2023}.
We apply two potential fields -- passing and scoring -- to estimate the chance of a successful pass and the chance of a goal.
The passing field includes functions such as opponent distance along the pass path and the distance between the ball and the receiver.
The scoring field uses the opening angle toward the opponent’s goal and the distance to it. For each teammate except the goalkeeper, both probabilities are multiplied, a hysteresis based on the prior choice is added, and the top option becomes the target only if direct scoring is less likely.
An example appears in \cref{f:passing}.
The final kick is refined because opponents often block direct passes. To lower interception risk, we use a stronger kick when the teammate need not move far toward the trajectory, shortening the time the ball rolls outside our control.

\section{Results}
\label{s:results}

In 2026, B-Human secured every soccer event it entered (\cf~\cref{t:results}). Across these matches, the team achieved the highest goals per minute (such as 1.24 at the RoboCup) and the fewest penalties per game (such as 0.4 at the RoboCup) of all participating teams. These results occurred on fields ranging from \qty{9}{\meter}~$\times$~\qty{6}{\meter} to \qty{22}{\meter}~$\times$~\qty{14}{\meter} with different layouts in 3~vs.~3 and 5~vs.~5 formats, typically lasting \qty{20}{\minute} but \qty{30}{\minute} at the World Humanoid Robot Games.

Two elements account for this outcome. First, B-Human had already built advanced software for the NAO robots, featuring complex team play and passing that scales well to different setups. Second, we developed new walk and kick skills, which made the existing software effective. Our walk is fast and robust, reaching up to \qty{2}{\meter\per\second} and \qty{120}{\degree\per\second}, enabling us to challenge ball possession across the field and in every one-on-one situation. We also focused on rapid and accurate kicks, with varied strengths to support dribbling, passing, and shots from any position.

\begin{table}[t]
    \caption{B-Human's game results in HSL Middle Division competitions in 2026}
    \label{t:results}
    \subfloat[RoboCup German Open]
    {
        \resizebox{0.77\width}{!}{%
            \begin{tabular}{r|l|r@{$:$}l}
                Home & Away & \multicolumn{2}{c}{Score} \\\hline
                HTWK Robots & \textbf{B-Human} & 1 & \textbf{11} \\
                \textbf{B-Human} & whIRLwind Amsterdam & \textbf{10} & 0 \\
                \textbf{B-Human} & Berlin United & \textbf{10} & 0 \\
                HULKs & \textbf{B-Human} & 0 & \textbf{10} \\
                Ruhrbot Devils & \textbf{B-Human} & 0 & \textbf{10} \\
                \textbf{B-Human} & Ruhrbot Devils & \textbf{10} & 0 \\
                \textbf{B-Human} & HTWK Robots & \textbf{6} & 1
            \end{tabular}
        }
    }
    \hfill
    \rotatebox{-90}{\hspace{1mm}$\xrightarrow{\qquad\text{\normalsize Final}\qquad}$}
    \hfill
    \subfloat[RoboCup]
    {
        \resizebox{0.77\width}{!}{%
            \begin{tabular}{r|l|r@{$:$}l}
                &  & \multicolumn{2}{c}{\phantom{\textbf{24}:\textbf{21}}} \\[-\normalbaselineskip]
                Home & Away & \multicolumn{2}{c}{Score} \\\hline
                \textbf{B-Human} & Rhoban & \textbf{10} & 1 \\
                CAU Mountain\&Sea & \textbf{B-Human} & 0 & \textbf{9} \\
                \textbf{B-Human} & RoboRoos & \textbf{10} & 0 \\
                \textbf{B-Human} & HTWK Robots & \textbf{8} & 1 \\
                \textbf{B-Human} & whIRLwind Amsterdam & \textbf{11} & 1 \\
                \textbf{B-Human} & RedbackBots  & \textbf{10} & 0 \\
                \textbf{B-Human} & whIRLwind Amsterdam & \textbf{9} & 0 \\
                \textbf{B-Human} & HTWK Robots & \textbf{6} & 0
            \end{tabular}
        }
    }

    \subfloat[RCAP Beijing Masters]
    {
        \resizebox{0.77\width}{!}{%
            \begin{tabular}{r|l|r@{$:$}l}
                \phantom{Ruhrbot Devils} &  & \multicolumn{2}{c}{\phantom{\textbf{10}:\textbf{10}}} \\[-\normalbaselineskip]
                Home & Away & \multicolumn{2}{c}{Score} \\\hline
                \textbf{B-Human} & whIRLwind Amsterdam & \textbf{6} & 2 \\
                \textbf{B-Human} & Inha-United & \textbf{16} & 0 \\
                \textbf{B-Human} & Starkit & \textbf{18} & 0 \\
                \textbf{B-Human} & Badger Bots & \textbf{15} & 0 \\
                \textbf{B-Human} & HTWK Robots & \textbf{9} & 0
            \end{tabular}
        }
    }
    \hfill
    \rotatebox{-90}{\hspace{1mm}$\xrightarrow{\quad\text{\normalsize Final}\quad}$}
    \hfill
    \subfloat[World Humanoid Robot Games]
    {
        \resizebox{0.77\width}{!}{%
            \begin{tabular}{r|l|r@{$:$}l}
                & \phantom{whIRLwind Amsterdam} & \multicolumn{2}{c}{} \\[-\normalbaselineskip]
                Home & Away & \multicolumn{2}{c}{Score} \\\hline
                CAU Mountain\&Sea & \textbf{B-Human} & 1 & \textbf{21} \\
                Aurora & \textbf{B-Human} & 1 & \textbf{21} \\
                \textbf{B-Human} & HyperYe & \textbf{24} & 0 \\
                \textbf{B-Human} & HTWK Robots & \textbf{13} & 3
            \end{tabular}
        }
    }
    \hfill
\end{table}

\section{Conclusion and Future Work}
In this paper, we provided a brief overview of our approaches to address the major challenges of competing with a new robot platform in the new RoboCup Humanoid Soccer League, taking into account the solutions we have already successfully deployed in the Standard Platform League. In addition, this paper contains the first comprehensive description of our team behavior approach, which enabled us not only to win all of our Middle Division games in 3~vs.~3 and 5~vs.~5 matches but is also capable of scaling up towards 11~vs.~11. This was demonstrated in the world's first humanoid 11~vs.~11 matches.

The main goal this year was to get everything up and running and to address the issues that had to be solved in order to perform well. Like many other teams, we had less than six months to work with the robots. We succeeded in this regard: all of our key objectives were met. In 2027, we can build upon this foundation. Future work will include, but is not limited to, more sophisticated 11~vs.~11 behaviors and, first and foremost, tracking and playing flying balls.

\bibliographystyle{splncs04}
\bibliography{bibliography.bib}

\end{document}